\documentclass[10pt,twocolumn,letterpaper]{article}

\usepackage{iccv}
\usepackage{times}
\usepackage{epsfig}
\usepackage{graphicx}
\usepackage{amsmath}
\usepackage{amssymb}
\usepackage[accsupp]{axessibility}

\usepackage{tabularx}
\usepackage{booktabs}

\usepackage{color}

\makeatletter
\@namedef{ver@everyshi.sty}{}
\makeatother
\usepackage{changes}

\usepackage[breaklinks=true,bookmarks=false]{hyperref}

\iccvfinalcopy 

\def\iccvPaperID{9418} 

\ificcvfinal\fi

\begin{document}

\title{Neural Photofit: Gaze-based Mental Image Reconstruction}

\author{Florian Strohm\textsuperscript{1}, Ekta Sood\textsuperscript{1}, Sven Mayer\textsuperscript{2}, Philipp Müller\textsuperscript{3}, Mihai B\^{a}ce\textsuperscript{1}, Andreas Bulling\textsuperscript{1}\\[0.25cm]
University of Stuttgart
  \texttt{\small\{florian.strohm,ekta.sood,mihai.bace,andreas.bulling\}@vis.uni-stuttgart.de} \\
LMU Munich
  \texttt{\small info@sven-mayer.com} \\
  German Research Center for Artificial Intelligence (DFKI)
  \texttt{\small philipp.mueller@dfki.de}
}

\maketitle
\ificcvfinal\thispagestyle{empty}\fi

\begin{abstract}
We propose a novel method that leverages human fixations to visually decode the image a person has in mind into a photofit (facial composite).
Our method combines three neural networks: An encoder, a scoring network, and a decoder.
The encoder extracts image features and predicts a neural activation map for each face looked at by a human observer.
A neural scoring network compares the human and neural attention and predicts a relevance score for each extracted image feature.
Finally, image features are aggregated into a single feature vector as a linear combination of all features weighted by relevance which a decoder decodes into the final photofit.
We train the neural scoring network on a novel dataset containing gaze data of 19 participants looking at collages of synthetic faces.
We show that our method significantly outperforms a mean baseline predictor and report on a human study that shows that we can decode photofits that are visually plausible and close to the observer's mental image.
\end{abstract}

\section{Introduction}

\begin{figure}[ht!]
    \centering
    \includegraphics[width=.9\linewidth]{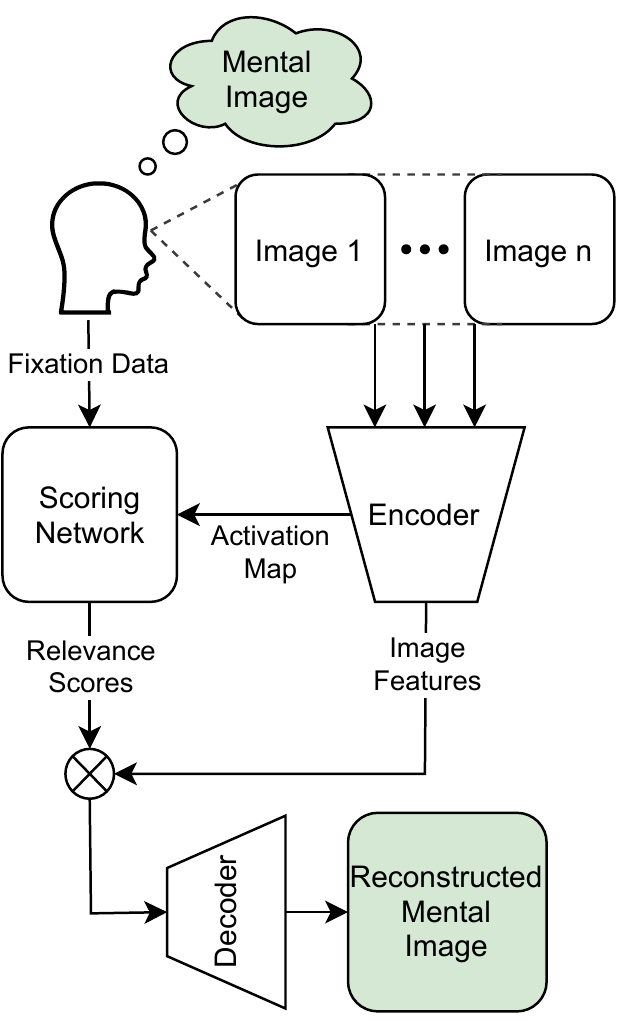}
    \caption{Overview of our method for gaze-based mental image reconstruction. With an image in mind, users search for similar facial features in multiple auxiliary images while their gaze is being recorded. An encoder extracts image features and corresponding neural activation maps from these images. A scoring network predicts the relevance of each image feature by comparing the fixation and neural activation maps. Image features are finally aggregated and decoded into a photofit.}
    \label{fig:overall_architecture}
\end{figure}

Visually decoding images that only exist in peoples' minds (also known as mental image reconstruction, MIR) has recently started to attract increasing research interest across a range of disciplines, including computational neuroscience, computational biology, and computer vision.
MIR is profoundly challenging given that the information required to succeed in this task is encoded in complex neural dynamics in the brain and not easily accessible from the outside.
The dominant approach for MIR has been to reconstruct mental images directly from brain activity recorded using functional magnetic resonance imaging~\cite{beliy2019voxels,guccluturk2017reconstructing,lin2019dcnn,shen2019deep,vanrullen2019reconstructing} or electroencephalography~\cite{date2019deep,shatek2019decoding}.

Another recent line of work has explored other sensing modalities, in particular human eye fixations.
Although fixations, in comparison, only provide an indirect measure of a person's mental image and as such complicate the reconstruction task further, they are promising because they are a less obtrusive and more practically useful measure of visual and cognitive processing, e.g., during scene perception~\cite{henderson2003human} or visual search~\cite{zelinsky1997eye}.
While several methods have been proposed to predict the target of visual search from fixations and image features~\cite{barz2020visual,borji2015eyes,sattar2015prediction,stauden2018visual,zelinsky2013eye}, only two previous works have tried to also visually decode (reconstruct) the search target~\cite{sattar2017predicting,sattar2020deep}.
These works first predicted the object class and attributes of the mental image from human eye fixations and then synthesised \textit{random samples} from the predicted class.

We significantly go beyond this state of the art by proposing a method that -- for the first time -- reconstructs the \textit{specific instance} of the mental image from eye fixations and auxilary images, without the system having seen this instance before.
We specifically focus on reconstructing facial images given this has high practical value also beyond criminology and is challenging given the large amount of facial appearance details.
Our method encodes multiple facial images looked at by an observer into separate feature vectors using a Siamese CNN encoder, fuses these vectors into a single feature vector using a novel scoring network, and finally decodes the mental image from this representation.
The scoring network compares neural activation maps for each output feature of the encoder with human fixations to predict how important each feature is for reconstructing the mental image.
Our method addresses the scarcity of gaze data at training time by allowing training of the encoder and decoder on large image datasets and only requiring joint image and gaze data for training the scoring network.

The specific contributions of our work are threefold:
First, we introduce an annotated dataset of human fixations on synthesised face images during face recognition that lends itself to studying the task of gaze-based mental image reconstruction.
Second, we introduce a novel problem formulation and method that, for the first time, allows us to synthesise a photofit -- that is, a visual reconstruction of the mental image of a face -- from human eye fixations.
Third, using this dataset as well as through a human study, we report on a series of experiments successfully demonstrating gaze-based reconstruction of mental face images\footnote{Code and other supporting material can be found at \url{https://perceptualui.org/publications/strohm21_iccv/}}.

\section{Related Work}

We review prior work on the related task of visual search target prediction and on reconstructing mental images from brain activity and eye gaze.
        
\subsection{Visual Search Target Prediction}

In a seminal work, Yarbus showed that gaze behaviour of observers reflects their visual task when looking at an image~\cite{yarbus1967eye}.
Later, Wolfe proposed an influential model for visual search behaviour that combined a parallel stage for processing information about basic visual features with a limited-capacity stage for more complex operations, such as face recognition, reading, or object identification~\cite{wolfe1994guided}.
Advances in machine learning have spurred interest in predicting the target of visual search.
That is, identifying the specific instance an observer is looking for among a set of potential target entities known a priori.
Pioneering work by Sattar et al. on natural images proposed a bag of visual words approach to search target prediction in an open-world setting~\cite{sattar2015prediction}.
Stauden et al. improved their method by using a pre-trained CNN for feature extraction~\cite{stauden2018visual}.
Similarly, Barz et al. used a pre-trained SegNet for encoding the fixation sequence and an SVN to predict the class of the image segment surrounding the likely search target~\cite{barz2020visual}.
In later work, Sattar et al. proposed the first method to predict the target class and attributes instead of only the target instance~\cite{sattar2017predicting}.
Work by Fang and Geman focused on an interactive setting in which faces where iteratively shown to the user and refined based on their feedback~\cite{fang2005experiments}.
Finally, Wang et al. studied a setting in which target prediction was performed after showing the stimulus~\cite{wang2019mental}.

\subsection{Mental Image Reconstruction}

Mental image reconstruction is the significantly more challenging task of not only predicting but visually decoding a target that only resides in an observer's mind and, as such, is also unknown to the system a priori.
Several works have focused on reconstructing mental images from brain activity, particularly using electroencephalography (EEG) and functional magnetic resonance imaging (fMRI).
While early work has, for example, relied on Gaussian mixture models~\cite{schoenmakers2014gaussian}, advances in deep learning methods have significantly improved reconstruction quality~\cite{shatek2019decoding,zhang2019multi}.
G{\"u}{\c{c}}l{\"u}t{\"u}rk et al. used probabilistic inference with deep learning based on EEG  to reconstruct gender, skin colour, and facial features~\cite{guccluturk2017reconstructing}. 
To improve the signal-to-noise ratio and thereby reconstruction quality, Date et al. instead used electrocorticography (ECoG) -- an invasive method in which electrodes are directly attached to the human brain -- in combination with a conditional generative adversarial network~\cite{date2019deep}.
In a parallel line of work, fMRI-based reconstruction methods have been developed and improved first using deep learning~\cite{guccluturk2017reconstructing, shen2019deep} and most recently by using generative adversarial networks (GANs)~\cite{dado2020hyperrealistic, lin2019dcnn, mozafari2020reconstructing, seeliger2018generative}, encoder-decoder architectures~\cite{beliy2019voxels, vanrullen2019reconstructing}, and a combination of both~\cite{ren2019reconstructing} to address the scarcity problem of fMRI data.

Instead of a direct mapping from brain signals into pixel space, Zaltron et al.~\cite{zaltron2020cg} utilised iterative user feedback to traverse the latent space of a pre-trained GAN. While achieving promising results users have to provide explicit feedback which can be demanding and difficult to provide. Furthermore, the coarse feedback necessitates many iterations to reach convergence. 

Only one previous attempt has been made at the even more challenging problem of reconstructing mental images from human gaze.
Sattar et al. proposed a Gaze Pooling Layer that integrated fixation information and a pre-trained deep convolutional image encoding into a semantic representation~\cite{sattar2020deep}.
This representation was then used with a Conditional Variational Auto-Encoder (CVAE) to visually decode \textit{random instances} of the predicted category and attributes of the search target~\cite{sattar2017predicting,sattar2020deep}.
In stark contrast, our method is the first to visually decode the \textit{specific instance} of the mental image from eye fixations.
\section{Data Collection}
\label{sec_data}

There currently does not exist a dataset for mental image reconstruction from eye fixations.
Sattar et al. \cite{sattar2015prediction} have released eye tracking data of participants performing visual search but that data is not suitable for the reconstruction task:
For one, they used image collages consisting of at least 20 images and, thus, the dataset lacks fine-grained eye tracking data within individual images that is required for a detailed reconstruction of a specific mental image.
Second, their use of natural images makes it difficult to study this novel task in a controlled fashion as well as -- given that perceptual metrics is an open research problem -- to quantify the reconstruction quality properly.
We therefore collected our own dataset and, as our first step, opted for a controllable image domain that allowed us to unveil the feasibility of gaze-based mental image reconstruction.

Faces are an ideal starting point for studying the underexplored mental image reconstruction task from fixations because there exist tools to generate faces while systematically controlling their appearance.
The software used in our data collection is FaceMaker~\cite{schwind2017facemaker} -- a tool that allows creating human-like faces by manipulating key facial features in a controlled fashion with 30 different sliders, such as eyebrow shape, skin colour, or width of the mouth (see supplementary material for a list of all sliders and example usage of FaceMaker).
To collect the dataset, we designed an eye tracking study that involved participants in creating and ranking face images using FaceMaker.
Participants designed their own target face based on a real face image to encode the appearance of this face in their memory. 

\begin{figure}[t]
  \centering
  \includegraphics[width=\linewidth]{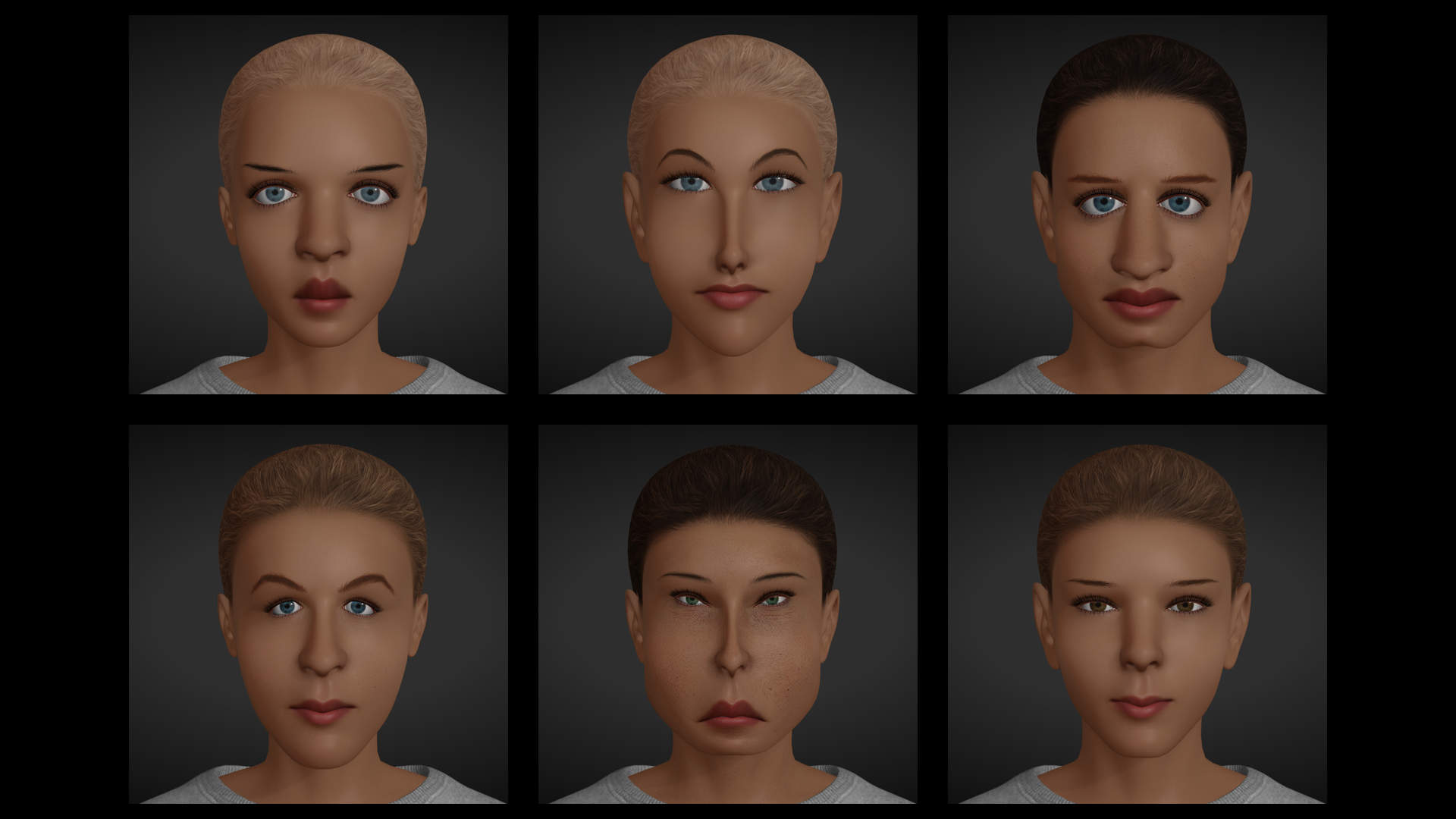}
  \caption{Sample image collage that was shown to participants in our data collection study. Two of the faces are fully random, while the other four contain the eyes, the nose, the mouth and the jaw of the participant's mental image (i.e. target face).}
  \label{fig:example_stimuli}
\end{figure}

\subsection{Participants and Apparatus}

We recorded gaze data of 19 participants (seven female) aged between 20 and 33 years (M=25.8, SD=3.4) whom we recruited through university mailing lists. 
All participants had normal or corrected-to-normal vision.
We used a stationary EyeLink 1000 Plus eye tracker that provides binocular gaze estimates at 2,000Hz.
Following best practices in eye tracking, we used a chin rest mounted in front of the participants to increase gaze estimation accuracy.
The face stimuli were shown on a 24.4-inch screen with a resolution of $1920\times1080$ pixels that was placed 90cm in front of the participants.
Each face was $14.3$cm$\times 14.3$cm in size corresponding to $8.9^{\circ}$ degrees of visual angle.

\subsection{Procedure}

After giving their consent and completing a short demographics questionnaire, we asked participants to sit on the chair and place their head in the chin rest.
We then guided them through the following process twice -- once with a female and once with a male target face (random order).
We first showed them a real face from the celebA dataset \cite{liu2015faceattributes} that we handpicked for diversity (no image was picked twice).
Next, we asked them to recreate the face using FaceMaker \cite{schwind2017facemaker} within five minutes.
The resulting image was the \textit{Target Face}.
Based on the \textit{Target Face}, we generated eight sets of six images each that participants had to compare to the target in eight trials.
Each trial started with an eye tracker calibration and validation. 
Afterwards, we showed the \textit{Target Face} again for ten seconds, asking participants to memorise it.
Then, participants were shown the six generated images and they had 30 seconds to rank them (see \autoref{fig:example_stimuli} for an example stimulus).
Participants were still able to finish their ranking after 30 seconds without seeing the stimulus.
Then the next trial started.
After finishing the eight trials, participants were shown a new celebA image to recreate and repeate the previous process. 

We created the six images based on the following procedure. We generated six fully random FaceMaker images, but set the image features \textit{Gender} and \textit{Skin Colour} to the \textit{Target Image} generated by the participant. These specific features could easily be defined by a user before MIR and, therefore, do not need to be reconstructed using fixations.
For four images, we set the image features to represent once the \textit{eyes}, once the \textit{nose}, once the \textit{mouth} and once the \textit{jaw} of the participants' \textit{Target Image}. See supplementary material for detailed information about which FaceMaker slider was assigned to these four groups. Summarising, we created six \textit{Auxiliary Images} shown to participants which always consisted out of two fully random faces and four images representing the \textit{eyes}, \textit{nose}, \textit{mouth}, or \textit{jaw} of the \textit{Target Image}.
We used these four facial regions because they are most important for face recognition~\cite{davies1977cue,ellis1979identification,fraser1990reaction,sinha2006face,young1985matching}.
\section{Gaze-based Mental Image Reconstruction}

The task of gaze-based mental image reconstruction involves learning a mapping $\{(I_i,G_i)|i=1...n\} \mapsto I_M$, i.e., 
given a set of auxiliary images $\{I_i|i=1...n\}$ observed by a human yielding a set of fixations for each image $\{G_i|i=1...n\}$, reconstruct the mental image $I_M$ the observer had in mind.
Training a generative model for mental image reconstruction end-to-end would require large amounts of joint fixation and image data $\{(I_i,G_i)|i=1...n\}$, which is impractical \cite{sattar2020deep}.
To overcome this, our method consists of three separately trained models: an encoder, a novel scoring network and a decoder.
This approach allows us to train the encoder and decoder networks solely on image data, while training of the scoring network only requires a small amount of joint image and fixation data.
\autoref{fig:overall_architecture} provides an overview of the architecture of our method.

\paragraph{Encoder.}

The encoder $e$ is trained to learn a mapping $e: I \mapsto F$, where $I$ is an image from the same domain as the mental image, and $F$ are features extracted from $I$.
$I$ is passed through several convolution layers to extract spatial features that are reduced to a vector using a global average pooling (GAP) layer~\cite{lin2013network}.
This vector is used to predict a set of low dimensional generative parameters defining the image.
These parameters depend on the specific MIR setting used.
In addition to image features, the encoder produces neural activation maps for each output feature that can be interpreted as the attention of the encoder over the input image.
Following Zhou et al.~\cite{zhou2016learning}, the neural activation map $M^a_f$  for a feature $f \in F$ is given by:
\[
M^a_f(x,y) = \sum_k w^f_k \cdot f_k(x,y), \tag{1} \label{eq:activation_map}
\]
where $w^f_k$ is the weight between neuron $f$ of the output layer and activation $k$ of the GAP layer, and $f_k(x,y)$ is the activation of the $k$-th convolution kernel in the last convolution layer at location $(x,y)$.
For this method to work, the second to last layer should be a GAP layer, and the spatial resolution of the last convolution layer defines the resolution of the activation maps. 
Therefore, the encoder is optimised to extract meaningful image features while being able to calculate activation maps with a sufficiently large spatial resolution.
It is used to extract features and corresponding activation maps for each image $\{I_i|i=1...n\}$.

\paragraph{Scoring Network.}

\begin{figure}[]
    \centering
    \includegraphics[width=\linewidth]{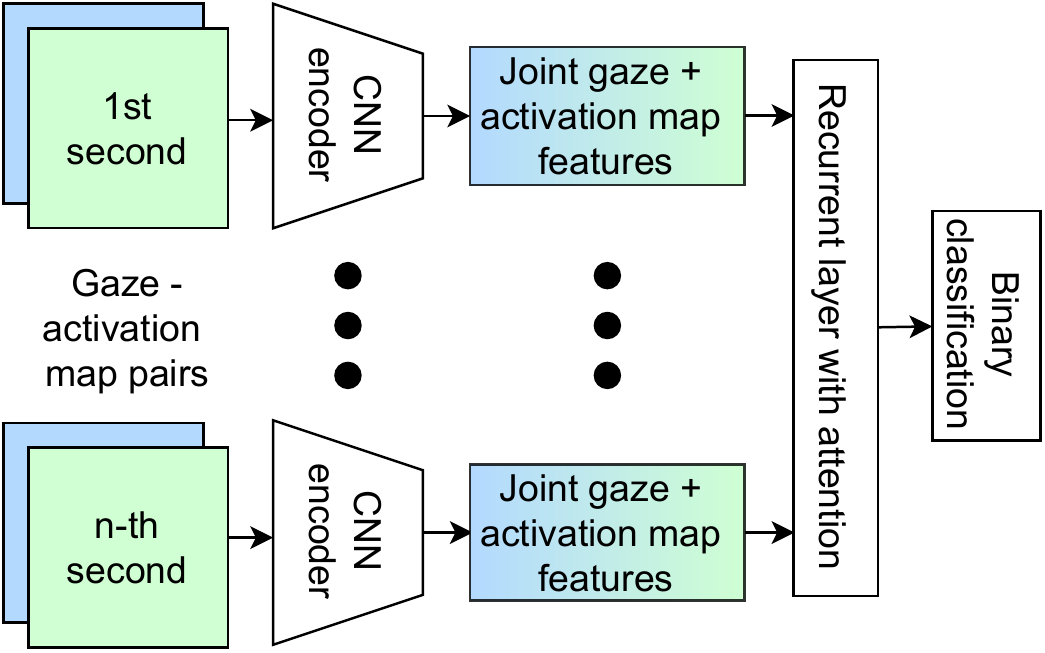}
    \caption{Architecture of our scoring network. Pairs of fixation and activation maps over time are encoded into a joint feature space. A subsequent recurrent layer with attention extracts time dependent features which are used to predict whether the image feature corresponding to the activation map is relevant for the mental image.}
    \label{fig:architecture}
\end{figure}

To reconstruct a mental image, we have to visually combine a set of image features $\{F_i|i=1...n\}$ from the set of images $\{I_i|i=1...n\}$ observed by the user.
The core idea of our method is a scoring network that compares ground-truth human fixations with trained neural attention and predicts a score that indicates the relevance of image features for the reconstruction (cf. \autoref{fig:architecture}).
The scoring network takes a fixation map $M^g_i$ and an activation map $M^a_{i,f}$ as input and predicts whether the image feature $f \in F$ of image $I_i$ is relevant for the mental image $I_M$, $P(f_{\text{relevant}} | M^g_i, M^a_{i,f})$.
While the activation map is predicted by the encoder, the fixation map is created by placing 2D Gaussians at each fixation location weighted by the duration.
Similar to multi-duration saliency~\cite{fosco2020much}, instead of creating one fixation map for each image using all fixations of a trial, we create multiple fixation maps over time, as shown in \autoref{fig:architecture} (left).
A Siamese CNN encoder is applied to each of the input tuples and extracts joint fixation and activation map features. 
The encoder consists of multiple convolution layers combining the information of the fixation and activation maps and extract spatial feature maps.
A GAP layer combines the feature maps into a feature vector, resulting in one vector for each time step in the input sequence.
These feature vectors are passed into a recurrent layer with an attention mechanism, which
enables the model to focus on the most important time steps of the input sequence.
The resulting feature vector is used in the output layer for binary classification.
The features of the mental image $F_M$ are then reconstructed as follows:
\[
F_M = \frac{\sum_{i=1}^n e(I_i) \cdot \text{score}(e(I_i),G_i)}{\sum_{i=1}^n \text{score}(e(I_i),G_i)}. \tag{3} \label{eq:scoring_funct}
\]
That is, the final mental image features are a linear combination of the features of images shown to a user, with the normalised predicted scores as the coefficients.

\paragraph{Decoder.}

The decoder $d$ is trained to learn a mapping $d: F_M \mapsto I_M$, where $F_M$ are combined image features given by the scoring network and $I_M$ is the mental image.
The input features $F_M$ are reshaped into a tensor, which is subsequently passed through transposed convolution layers to increase the resolution.
A final transposed convolution layer with three kernels, one for each image channel, produces the output image $I_M$.
\section{Experiments}

\definecolor{myGreen}{HTML}{006600}
\definecolor{myOrange}{HTML}{ff8000}
\definecolor{myRed}{HTML}{FF0000}

\begin{figure*}[t]
    \centering
    \includegraphics[width=\linewidth]{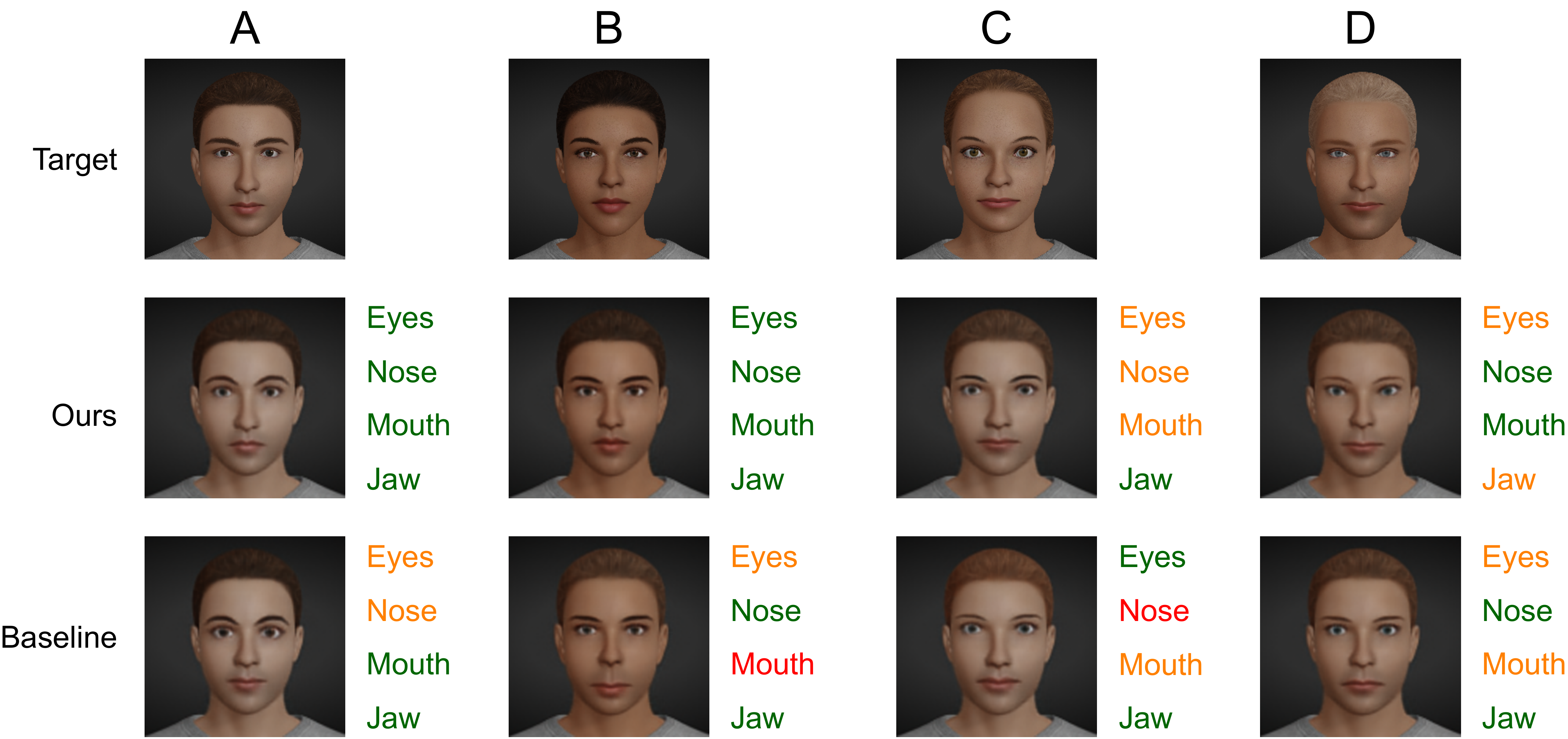}
    \caption{Sample images reconstructed by our method (Ours) in comparison with the respective target and baseline.
    The colour-coded labels for the different facial regions indicate reconstruction quality (\textcolor{myGreen}{high}, \textcolor{myOrange}{medium}, or \textcolor{myRed}{low}). We calculated these by assigning each feature group to one of three equidistant bins according to their mean absolute slider distance (MASD). Columns A and B show two of the best, while C and D show two of our worst mental image reconstructions.}
    \label{fig:reconstructions}
\end{figure*}

\subsection{Model Training}

To train the encoder and decoder, we generated 100K face images with a resolution of $128 \times 128$ pixels using the Facemaker synthesis software~\cite{schwind2017facemaker}.
To this end, we randomly sampled each slider value from a uniform distribution.
We generated an additional 30K images for validation to determine the best hyper-parameters for our models.
Both models were trained to minimise the mean squared error loss using the Adam optimiser~\cite{kingma2014adam} with default parameters and a batch size of 32.

To train the scoring network we split the collected fixation-image data into a training, validation, and test set.
We randomised the trials of each participant and selected 12 trials for training, two for validation, and two for testing. 
One participant finished only six trials, of which we used five for training and one for validation.
In total, this resulted in 221 trials for training, 37 trials for validation, and 36 trials for testing.
Given that we know for each trial which face contains relevant image features, together with the trained encoder, we created the fixation and neural activation maps for each face and feature. 
The ground-truth was set to one if the feature corresponded to the mental image and zero otherwise.
Because most of the features are not relevant for the mental image, this resulted in a class imbalance of $5:1$.
We therefore augmented the training set by flipping the fixation and activation maps, individually and jointly.
This resulted in a more balanced ratio of $5:4$, providing us with about 45K train, 5K validation, and 5K test samples.
We used the Adam optimiser with default parameters~\cite{kingma2014adam} for training using a binary cross-entropy loss and a batch size of 32. 

\subsection{Implementation Details}

\paragraph{Encoder.}

The input image resolution for the encoder was $128\times128$ pixels with RGB channels.
It was passed through four convolution layers with 32, 64, 128, and 256 kernels of size $4\times4$.
After each convolution layer, we applied a ReLU activation function and batch normalisation.
The first and third convolution layer convolved the kernels with a stride of one in each dimension, while the second and fourth layer used a stride of two.
Therefore, the feature maps of the last convolution layer were of dimensionality $32\times32\times256$ and the spatial resolution of the activation maps was $32\times32$.
See supplementary material for example activation maps of our model.
These feature maps were reduced to a 256-dimensional vector using global average pooling (GAP) layer~\cite{lin2013network}.
In this work, the generative parameters predicted by the encoder correspond to the thirty slider values from Facemaker.
Our final encoder model has a total of about 700K trainable parameters.
 
\paragraph{Scoring Network.}

We used a temporal resolution of one second for our final model.
Given that the trials of our dataset had a duration of 30 seconds, we input 30 fixation-activation map pairs into the scoring network.
The encoder within the scoring network consisted of a 3D convolution layer with 10 kernels of size $4\times4\times2$ to combine the information of the fixation and activation maps.
This was followed by two 2D convolution layers with 14 and 16 kernels of size $4\times4$ to further refine the features as well as a ReLU activation and batch normalisation.
A GAP layer combined the extracted spatial features and yielded a total of 16 features per time step. 
Each of the resulting 16-dimensional feature vectors was the input to a single GRU layer~\cite{cho2014properties} with 30 hidden units and an attention mechanism, as described by Zhou et al.~\cite{zhou2016attention}.
The output of this layer was a 30-dimensional feature vector used in a final dense layer for binary classification.
In total, the scoring network consisted of about 10K trainable parameters.
 
\paragraph{Decoder.}

The input image features of the decoder were passed through a dense layer with $4\times4\times256$ neurons, followed by a ReLU activation, and batch normalisation.
The resulting vector was reshaped into a $4\times4\times256$ tensor and was subsequently passed through four transposed convolution layers.
The convolution layers had 128, 64, 32, 16 kernels of size $4\times4$, each with a stride of two in each dimension and were followed by a ReLU activation and batch normalisation.
A final transposed convolution layer with three kernels, one for each image channel, and Sigmoid activation decoded the output image $F_M$.
The final decoder model consisted of about 840K trainable parameters.
We optimised all model parameters on the validation set and, in the following, report results obtained on the left-out test set.

\subsection{Evaluation Metric}

To quantify the performance of our method, we define the mean absolute slider distance (MASD):
\[
\text{MASD} = \frac{1}{30} \sum_{i=1}^{30} |s^p_i-s^t_i|, \tag{4} \label{eq:MASD}
\]
where $s^p_i$ is the predicted value and $s^t_i$ is the target value for slider $i$.
By changing the scoring function defined in \autoref{eq:scoring_funct}, we can evaluate our method in terms of accuracy:
\[
F_{M_f} = \max\limits_{I_{i,f}} \text{score}(e(I_{i})_f, G_i). \tag{5} \label{eq:scoring_funct_class}
\]
Instead of defining $F_M$ as a linear combination of all features, we select each feature from the face achieving the highest score for $f$.
We report  micro-averages, i.e., we compute them over the features not the feature groups.
The results are compared to a mean baseline reconstruction that was generated by averaging the six auxiliary faces.
We experimented with more sophisticated baselines using a state-of-the-art landmark detector~\cite{bulat2017far} to segment faces and accumulated fixation durations on each segment for scoring.
Since the performance of this method was inferior we only report the mean baseline in the following.

\subsection{Reconstruction Results}

\begin{table*}[t]
\begin{center}
\begin{tabularx}{\textwidth}{l *5{>{\centering\arraybackslash}X}||*5{>{\centering\arraybackslash}X}}
\toprule
 & \multicolumn{5}{c||}{Accuracy} & \multicolumn{5}{c}{MASD}\\
\cmidrule(r){2-6}\cmidrule(l){7-11}
Model & Eyes & Nose & Mouth & Jaw & All & Eyes & Nose & Mouth & Jaw & All\\
\midrule
\textbf{Ours} & \textbf{77.0\%} & 45.3\% & \textbf{58.9\%} & \textbf{44.6\%} & \textbf{61.9\%} & \textbf{17.75} & \textbf{16.55} & \textbf{15.10} & \textbf{24.93}  & \textbf{23.37} \\
5s & 73.2\% & 42.6\% & 49.7\% & 38.5\% & 56.8\% & 21.14 & 18.92 & 17.82 & 26.13 & 25.15 \\
30s & 67.3\% & \textbf{54.7\%} & 53.5\% & 43.2\% & 57.9\% & 19.64 & 16.67 & 16.95 & 26.95 & 23.96 \\
NoAtt & 56.8\% & 50.7\% & 51.9\% & 42.6\% & 52.2\% & 25.87 & 19.25 & 19.65 & 25.06 & 26.35 \\
\midrule
Baseline & 16.7\% & 16.7\% & 16.7\% & 16.7\% & 16.7\% & 31.2 & 23.36 & 23.98 & 29.96 & 30.26 \\
\bottomrule
\end{tabularx}
\end{center}
\caption{Performance of our method, several ablated versions, and the baseline for reconstructing different facial regions in terms of accuracy and mean absolute slider distance (MASD).}
\label{tbl:results}
\end{table*}

\autoref{tbl:results} shows the performance of our method and different ablated versions in terms of MASD and accuracy.
As can be seen, our method achieves an average MASD of $23.37$ on the test set, which corresponds to an average error of $12.8\%$ compared to the \textit{baseline} model that achieves an average MASD of $30.26$ (average error of $16.6\%$).
The average accuracy of our method is $61.9\%$, which is significantly higher than chance level of $16.7\%$, i.e., selecting each feature from a stimulus at random.

The \textit{5s model} only uses six fixation maps instead of 30, each containing five seconds of fixation data.
This model performs worse overall which indicates that the higher temporal resolution of one second per fixation map is beneficial.
The \textit{30s model} only uses one fixation map containing all fixations within the 30 seconds trial.
Even without temporal information this model achieves a good overall performance.
Nevertheless, it performes worse in all cases except for nose-related features, underlining the benefit of leveraging the temporal characteristics inherent in gaze data. Even though our model achieves a lower accuracy for nose-related features, it achieve the lowest MASD. This indicates that, even if the highest relevance score is not assigned to the face containing the correct feature, a linear combination of that feature over all faces still results in a good reconstruction. To analyse the impact of the attention mechanism we excluded it in the \textit{NoAtt model}.
This model achieves the lowest overall performance, especially for the eyes, suggesting that the attention-induced bias helps the model to generalise and cope better with the little training data.

\begin{figure}[t]
    \centering
    \includegraphics[width=\linewidth]{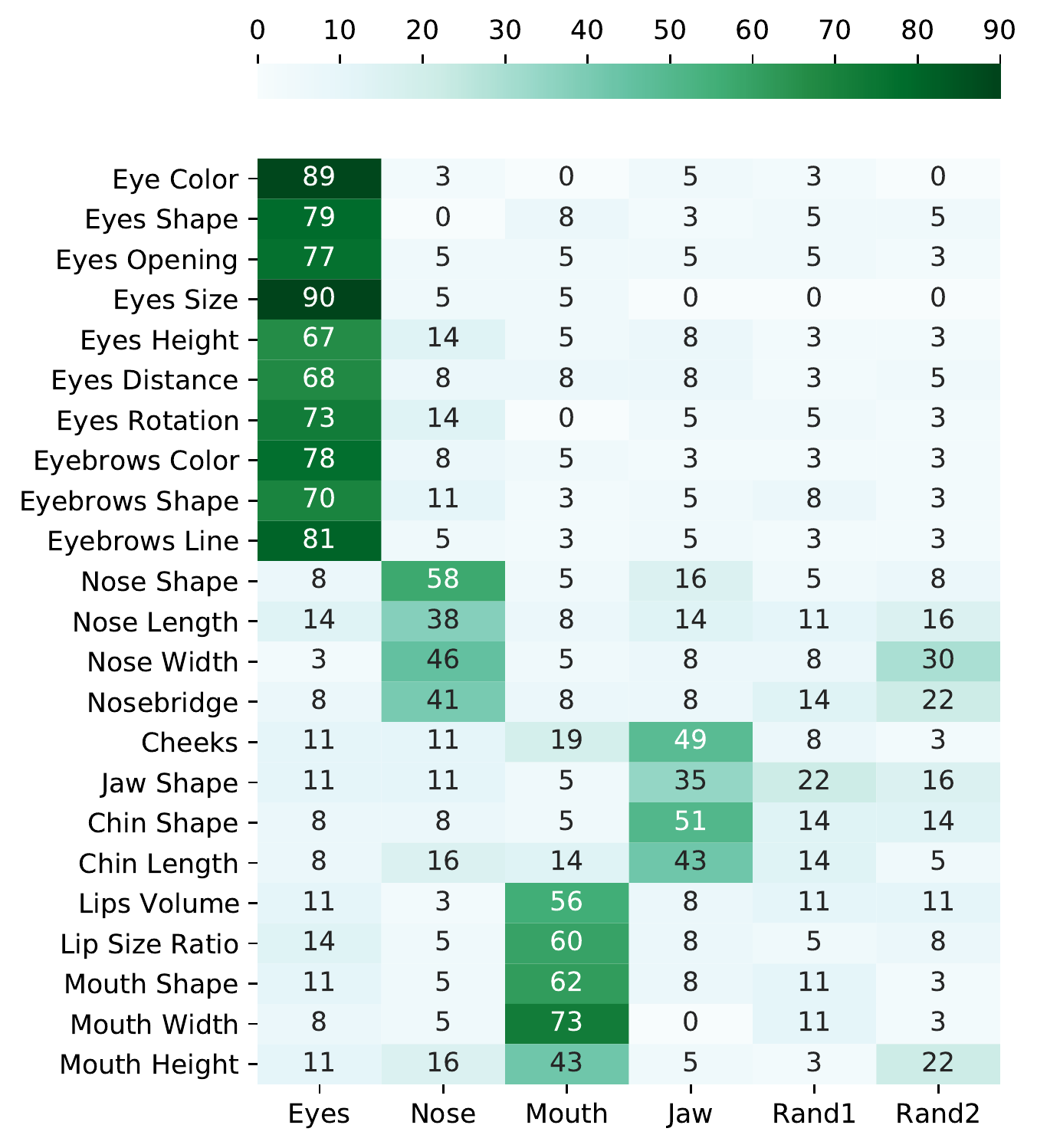}
    \caption{Accuracy of our method for different facial features using a modified scoring function to form a classification objective. Each cell $c_{i,j}$ indicates the percentage of cases in which feature $f_i$ was extracted from auxiliary image $I_j$. 
    }
    \label{fig:confusion_matrix}
\end{figure}

\autoref{fig:reconstructions} shows four sample target faces, reconstructions generated by our method, and baseline reconstructions as well as colour-coded indications of the reconstruction quality of the different facial regions (see supplementary material for all test set reconstructions).
We calculated these by assigning each feature group to one of three equidistant bins according to their MASD.
Finally, \autoref{fig:confusion_matrix} shows our model's accuracy with the modified scoring function to calculate accuracy for each image feature.
The value in each cell $c_{i,j}$ shows the percentage of how often the value for a feature $f_i$ was selected from auxiliary face $I_j$.

\subsection{User Study}
To assess the subjective quality of the reconstructions, we conducted a 24-participant user study.
In the study we showed participants three faces: a target face from the test set, the reconstruction from our method, and the face reconstructed using the baseline method.
For every face in the test set, we asked participants which of the two reconstructions they thought resembled the target face more closely. 
For a total of 36 faces in the test set, participants selected our faces $79\%$ of the time on average (chance level 50\%). If we consider the majority vote, the face reconstructed with our method was selected 32 out of 36 times.
\section{Discussion}

\paragraph{Reconstruction Performance.}

In this work, we introduced the task of reconstructing the mental image from human  fixations.
Our results showed that we could reconstruct the mental image significantly better than the baseline method, with accuracy ranging from 45\% to 77\% and a MASD ranging from about 25 to 15 (see \autoref{tbl:results}).
We further showed that our method performs better at reconstructing important facial features (such as the eyes, nose, mouth, and jaw) than the baseline and overall was able to generate plausible photofits that show high similarity to the mental image (see \autoref{fig:reconstructions}).
Although researchers have investigated the influence of mental image retrieval on visual search behaviour, it was unclear whether mental image reconstruction from fixations was possible \cite{pearson2008functional,pearson2015mental,moriya2018visual}.
As such, our results are promising and underline the effectiveness of the proposed method.
We also noticed, however, that our method often fails to reconstruct the correct hair colour.
We observed that participants rarely fixated on the hair of the auxiliary faces and hypothesise that they instead were able to identify this feature in their peripheral vision.

In the confusion matrix in \autoref{fig:confusion_matrix}, four groups can be identified, each corresponding to a set of sliders for the eyes, nose, jaw, and mouth regions.
Each slider's values were selected most often from the auxiliary face that contained the correct value.
There is no strong confusion between auxiliary faces for any feature; misclassifications are rather equally distributed across the faces.

\paragraph{Analysis of Feature Groups.}

\begin{figure}[t]
    \centering
    \includegraphics[width=\linewidth]{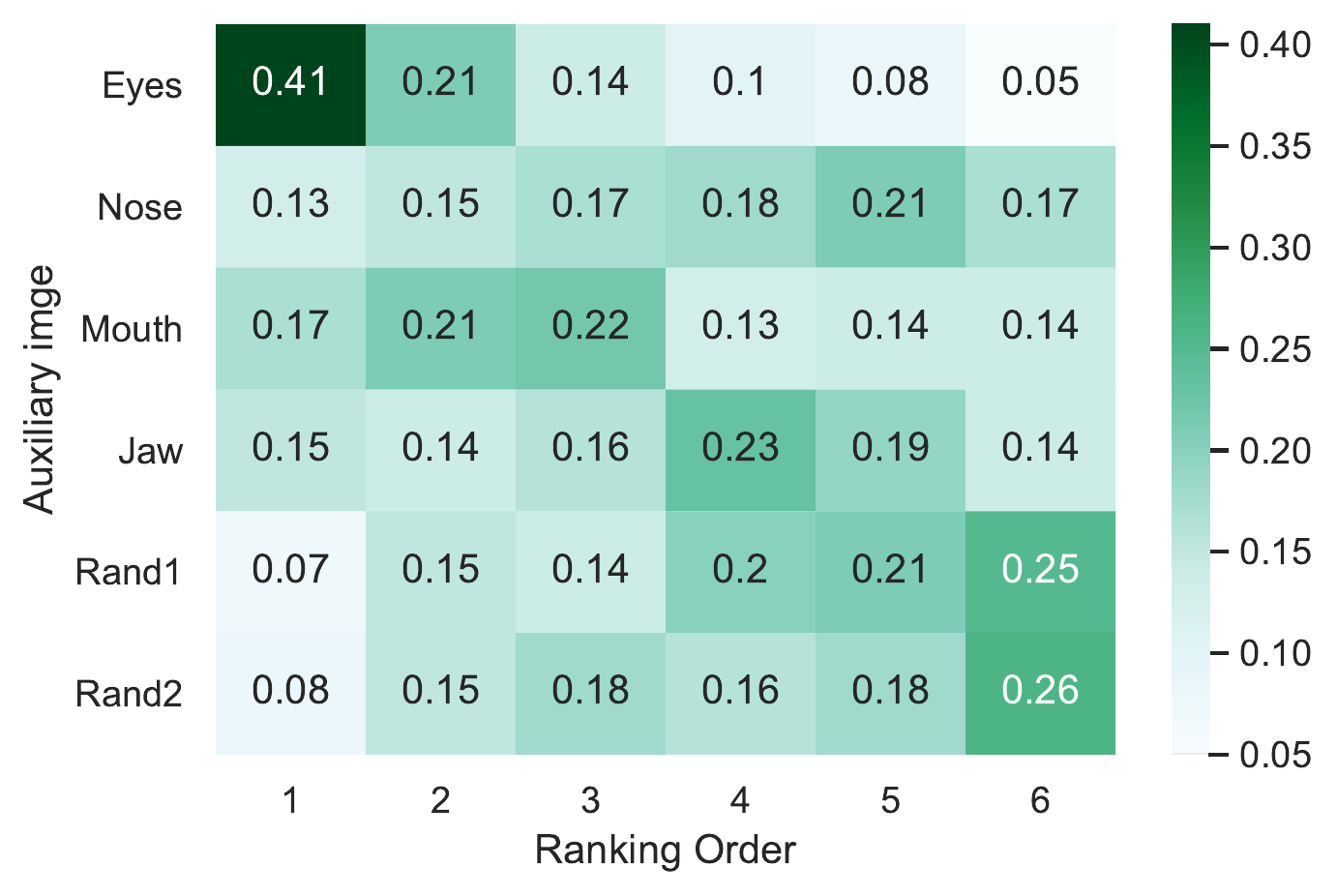}
    \caption{Percentage of ranks for each face given by the participants over the whole dataset.
    }
    \label{fig:ranking}
\end{figure}

\autoref{tbl:results} also shows that the accuracy of our method is the highest for eye-related features, followed by mouth, nose, and jaw.
This reflects the average rank participants assigned to auxiliary faces containing these features during data collection.
\autoref{fig:ranking} shows how often the participants assigned a specific rank to each image with labels on the y-axis indicating the type of auxiliary face.
The order of the average rank is in line with previous findings in psychology ~\cite{davies1977cue,ellis1979identification,fraser1990reaction,sinha2006face,young1985matching} showing that eyes are the most important feature for face recognition (average rank 2.34), followed by mouth (average rank 3.31) and nose (average rank 3.73).
Furthermore, we can observe that, as expected, the two distractor faces follow a similar rank distribution and are ranked the lowest on average (average ranks of 4.14 and 4.02, respectively).

\paragraph{Fixation Duration over the Auxiliary Faces.}

\begin{figure}[t]
    \centering
    \includegraphics[width=\linewidth]{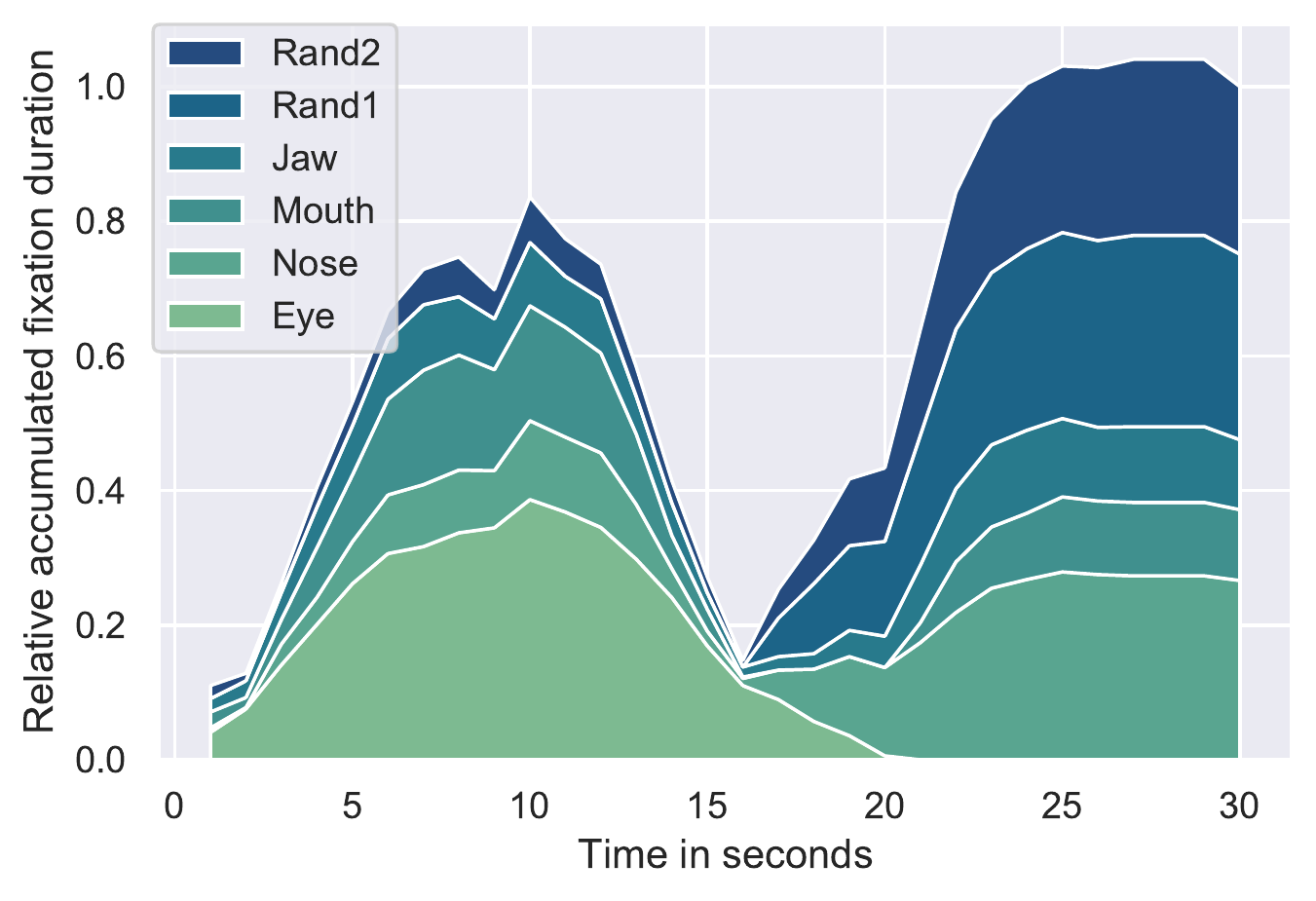}
    \caption{Accumulated fixation duration over time for the six auxiliary faces. For each time step we subtracted the lowest accumulated fixation to better highlight the change in attention over time. During the intial 15 seconds, there is more attention on images containing relevant features (Jaw, Mouth, Nose, Eye), while attention subsequently shifts to the random images (Rand1, Rand2).}
    \label{fig:gaze_over_time_diff}
\end{figure}

\autoref{fig:gaze_over_time_diff} shows a stacked graph representing the relative fixation duration for each face accumulated over time.
For each time step, we subtracted the graph with the lowest accumulated fixation duration from all graphs to better highlight changes in fixation distribution over time.
Interestingly, participants mostly focused on the images containing features relevant to the mental image within the first 15 seconds of viewing time.
Since participants attribute more attention to features known to be more relevant for identification, our model receives more information for these features to infer the relevance score of a feature. This attention distribution correlates with the average ranking of the faces and with the facial feature importance for face perception, as discussed above.
After around ten seconds, participants' attention shifts to increasingly focus on the distractor faces, which gain the highest attention during the last ten seconds of the trial. Since the distractor faces are ranked low on average, participants rank them later within a trial, resulting in higher fixation durations at that time.

\paragraph{Analysis of the Attention Mechanism.}

Our scoring network contains a recurrent layer with an attention mechanism (see \autoref{fig:architecture}), allowing us to analyse how the network distributes its attention over the temporal input sequence.
By extracting the average attention distribution over the test set, we observe that the model assigns a higher weight to features between the fifth and tenth second.
This window overlaps with the time period where participants focused on auxiliary images containing relevant features (see \autoref{fig:gaze_over_time_diff}).

\section{Conclusion}

In this work, we introduced the first method to visually reconstruct the facial image a person has in mind only from their eye fixations. 
Gaze-based mental image reconstruction is profoundly challenging given that gaze is only an indirect measure of mental imagery and subject to significant variability caused by parallel cognitive processing.
In addition, joint gaze and image data is scarce, preventing the application of existing large-scale methods.
In stark contrast, key components of our method can be trained solely on image data and it requires only a small amount of gaze-augmented image data.
A second key contribution of our work is to formulate the reconstruction as a similarity scoring task between human fixation and neural attention maps.
Through quantitative evaluation and a human study, we showed that our method significantly outperforms a baseline method and can generate photofits that are visually similar to the mental image.
These significant advances point the way to future methods that can reconstruct mental images from gaze in other domains, including real human faces.
\section{Acknowledgements}

F. Strohm and A. Bulling were funded by the European Research Council (ERC; grant agreement 801708).
E. Sood was funded by the Deutsche Forschungsgemeinschaft (DFG, German Research Foundation) under Germany's Excellence Strategy -- EXC 2075 -- 390740016.
M. B\^{a}ce was funded by a Swiss National Science Foundation (SNSF) Early Postdoc.Mobility Fellowship.
P. Müller was funded by the German Ministry for Education and
Research (BMBF) under grant number 01IS20075.

{\small
\bibliographystyle{ieee_fullname}
\typeout{} 
\bibliography{references}
}

\end{document}